\documentclass[a4paper,twoside]{article}

\usepackage{epsfig}
\usepackage{subcaption}
\usepackage{calc}
\usepackage{amssymb}
\usepackage{amstext}
\usepackage{amsmath}
\usepackage{amsthm}
\usepackage{multicol}
\usepackage{pslatex}
\usepackage{apalike}
\usepackage{tabularx}
\usepackage[bottom]{footmisc}
\usepackage{tikz}
\usetikzlibrary{quantikz}
\usepackage{adjustbox}
\usepackage{booktabs}
\usepackage{multirow}

\usepackage{todonotes}

\usepackage{SCITEPRESS}     

\begin{document}

\title{Compression of GPS Trajectories using Autoencoders}

\author{\authorname{
Michael Kölle\sup{1}, 
Steffen Illium\sup{1},
Carsten Hahn\sup{1},\\
Lorenz Schauer\sup{1},
Johannes Hutter\sup{1},
Claudia Linnhoff-Popien\sup{1} }
\affiliation{\sup{1}Institute of Informatics, LMU Munich, Oettingenstraße 67, Munich, Germany}
\email{\{michael.koelle\}@ifi.lmu.de}
}


\abstract{The ubiquitous availability of mobile devices capable of location tracking led to a significant rise in the collection of GPS data.
Several compression methods have been developed in order to reduce the amount of storage needed while keeping the important information. 
In this paper, we present an lstm-autoencoder based approach in order to compress and reconstruct GPS trajectories, which is evaluated on both a gaming and real-world dataset.
We consider various compression ratios and trajectory lengths. 
The performance is compared to other trajectory compression algorithms, i.e., Douglas-Peucker.
Overall, the results indicate that our approach outperforms Douglas-Peucker significantly in terms of the discrete Fr\'echet distance and dynamic time warping. 
Furthermore, by reconstructing every point lossy, the proposed methodology offers multiple advantages over traditional methods.
}

\onecolumn \maketitle \normalsize \setcounter{footnote}{0} 

\section{Introduction}
The rising popularity of modern mobile devices such as smartphones, tablets or wearables and their now ubiquitous availability led to a vast amount of collected data. 
Additionally, such devices can determine the user's current location and may offer position-tailored information and services.
However, resource (energy) limitations apply.
In contrast to energy saving, embedded systems, smartphones are typically at the upper hardware limit.
When dealing with GPS information, the data analysis is being outsourced to data centers, often. 
A common way to transmit the data from the mobile devices to central servers is via cellular or satellite networks.
According to Meratnia and de By \cite{Meratnia2004},
around \textit{100Mb} of storage size are needed, if just 400 objects collect GPS information every ten seconds for a single day. Considering large vehicle fleets or mobile applications with millions of users tracking objects for long time spans, it is obvious that there is a need to optimize the transmission and storage of trajectories through compression. 


In this paper, we investigate and evaluate the approach of autoencoders for compression and reconstruction of GPS trajectories.
For evaluation, we identify adequate distance metrics that measure the similarities between
trajectories to evaluate the method and compare it to existing handcrafted compression algorithms, such as Douglas-Peucker. 
As main contribution, we will answer the following research questions:\\
\textbf{1)} How well does an autoencoder model perform in trajectory
		compression and reconstruction compared to traditional line
		simplification methods?\\
\textbf{2)} How well does an autoencoder model perform for different
		compression ratios and different trajectory lengths?

The paper is structured as follows: Section \ref{sec:Related} holds related work.
Section \ref{sec:Methodology} presents our methodology.
Evaluation is done in Section \ref{sec:eval}, and finally, Section \ref{sec:FutureWork} concludes the paper. 
\vspace{-1.5em}

\section{Related Work}\label{sec:Related}
A common method to compress (2-dimensional) trajectories are line simplification algorithms, such as the \textit{Douglas-Peucker} \cite{Douglas1973}.
In contrast, GPS trajectories include time as an important third dimension.
To overcome this misconception, Meratnia and de By \cite{Meratnia2004} state that the propose a variant called \textit{Top-Down Time-Ratio (TD-TR)} using the synchronized Euclidean distance (SED) as error metric.
Beside line simplification, \textit{road network compression} is applied in literature. Kellaris et al. \cite{Kellaris2009} combine GPS trajectory compression with map-matching using TD-TR. Richter et al. \cite{Richter2012} exploit the knowledge about possible movements with the use of \textit{semantic trajectory compression} (STC). The COMPRESS framework, introduced by Han et al. \cite{Han2017}, relies on knowledge
about the restricted paths an object can move on. Chen et al. \cite{Chen2013} exploit the existence of repeated patterns and regularity in trajectories of human movement by extracting so-called pathlets from a given dataset. 

The main focus of autoencoder research in recent years has been on the compression of image data, such as \cite{Gregor2016}. Oord et al. \cite{Oord2016} introduced PixelRNN, a generative model allowing the lossless compression of images with state-of-the-art
performance. The approach is based on \cite{Theis2015} where two-dimensional LSTM networks are proposed to model the distributions of
images. Toderici et al. \cite{Toderici2015} use a combination of LSTM and CNN elements to compress
thumbnail images. It is suitable for small images but falls behind manually composed algorithms
for larger resolutions. This downside is partially overcome in the subsequent
work \cite{Toderici2016} where an average performance better than JPEG can be achieved. Masci et al. \cite{Masci2011} propose a deep autoencoder based on CNN. The introduced convolutional
autoencoder (CAE) is trained layer-wise and the resulting weights are used as a
pre-training for a classification network. Rippel and Bourdev (2007) improve on the
aforementioned methods and suggest a generative adversarial network  \cite{Rippel2017}. The authors claim to double the compression ratio compared to JPEG and WebP while reaching the same quality. Beside image compression, autoencoders are applied for different purposes. E.g. Testa and Rossi (2015) use an autoencoder to compress biomedical signals \cite{Testa2015}. 

Overall, there is only few research dealing with the compression of time series data using
autoencoders.There is a combination of LSTM and autoencoder to compress a seismic signal \cite{Hsu2017}.
Similarly, Hejrati et al.  compress electroencephalogram data with a multi-layer perceptron \cite{Hejrati2017}.
\vspace{-1.5em}

\section{Methodology}\label{sec:Methodology}
In this section, we introduce our methodology for trajectory compression. Since the
trajectory of a moving object is usually represented as a time series of
discrete positions annotated with a timestamp, an RNN is
used to build the autoencoder' model. More precisely, an LSTM network architecture is applied, due to its remarkable performance in classification and prediction tasks with time series data.

\subsection{Encoder}

\begin{figure}[htb!]
	\centering
	\includegraphics[width=0.48\textwidth]{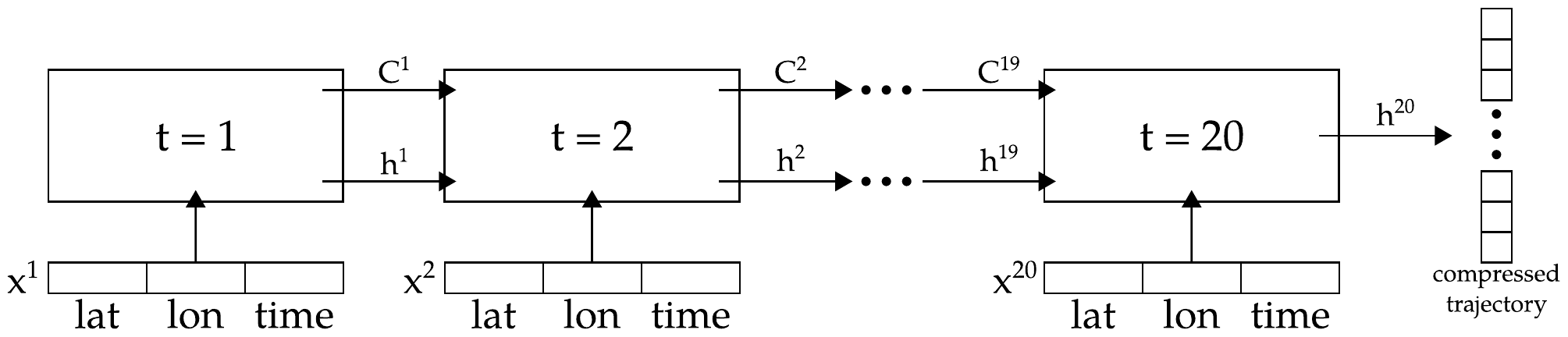}%
	\caption{Schematic of the LSTM encoder.}
	\label{fig:encoder}%
\end{figure}

As already stated, the encoder part transforms the input trajectory into a lower-dimensional latent representation.
At each time step a data point of the trajectory is fed to and processed by the LSTM. Depending on the used dataset, the three dimensions represent the position of an object inside a three-dimensional
space or a two-dimensional position and a timestamp, respectively. 
To reduce complexity, the length of the trajectory $|T|$ is fixed beforehand.
The output at every time step, but the last, is discarded as the final output contains the encoded information of the whole trajectory including the temporal dependencies. 
This latent representation forms together with the scaling and offset value from the
normalization step the compressed version of the original trajectory.
Per trajectory and feature two rescaling values for absolute position and scaling
factor are needed so that six values are already set for the compressed
trajectory. The number of values in the original trajectory are $|T| \cdot 3$,
since there are three values for each point in the trajectory. All the points
of the trajectories as well as the rescaling values and the latent variable of
the autoencoder are stored in 32 bit each, so that the number of values in the
original and compressed representation can be directly used to calculate the
compression ratio. Since the size of an input trajectory and the size of the
rescaling values are set, the compression ratio can be adjusted by modifying
the dimensionality of the encoder's output and therefore the latent variable
of the autoencoder. The compressed size is the sum of the numbers of latent and
rescale values.  Figure \ref{fig:encoder} depicts our LSTM encoder mapping a trajectory into the latent space.

\subsection{Decoder}
\begin{figure}[hpbt]
	\centering
	\includegraphics[width=0.48\textwidth]{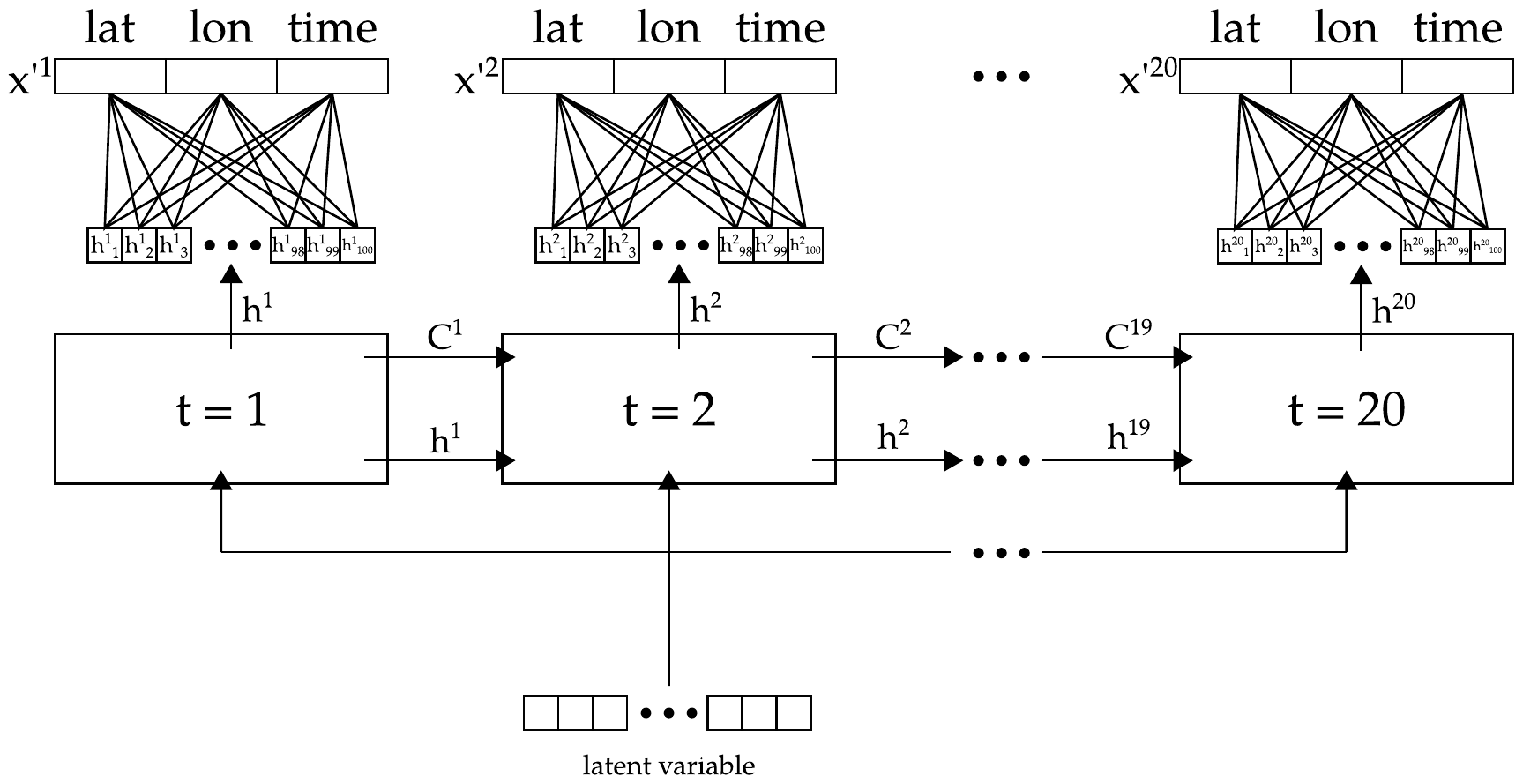}%
	\caption{Schematic of the LSTM decoder.}
	\label{fig:decoder}%
\end{figure}

Our decoder consists of a LSTM cell. At
each time step the latent variable is fed again to the network as its input and
the corresponding output represents the reconstructed three-dimensional data point.
During the training process the weights are optimized in order to produce the expected output and learn generalizable features of the
given data. Hence, the performance depends on the amount of weights which is dependent on the dimensionality of the input and output data. Since the weights are reused at each time step, the sequence length does not
influence the amount. Compared to other tasks, such as image compression or
natural language generation, the spatio-temporal data we use is low-dimensional, e.g., with a sequence length of 20 and a compression
ratio of 4 we deal with $\frac{20 \cdot 3}{4} = 15$ dimensions. Due to $2\times 3$ variables for offset and scaling factor from the normalization step, the latent variable is of 15 - 6 = 9 dimensions. Hence, the input size of the decoder at each time step is 9, while the output size is 3. The number of weights $n_{W}$ available in the cell depends on the amount of neural
layers. In a LSTM cell, gates are implemented as neural
layers and the block input as well as the output is activated by a layer. The
number of weights can therefore be calculated by:
$$
n_{W} =
4 \cdot ((n_{I} + 1) \cdot n_{O} + {n_{O}}^2)
$$
where $n_{I}$ is the input, and $n_{O}$ the output dimensionality.
The additional 1 is due to the bias term. The factor 4 represents the
three gates of a LSTM and the input activation. In the example, the decoder would have 156 weights available to fit to the data. This is insufficient for a
complex task, such as trajectory reconstruction. Hence, we set the output dimensionality of the decoder to 100 in order to deal with enough parameters that can be
adjusted to fit the temporal dependencies between points. In order to reduce the output back to a three dimensional space, a fully connected layer with three output neurons is added at the LSTM
output. Figure \ref{fig:decoder} depicts our architecture of the decoder.

Sutskever et al. \cite{Sutskever2014} stated that the performance of Sequence to Sequence
Autoencoders can be improved significantly, if the input sequence is reversed
before it is fed to the network. The target sequence stays in the original
order nonetheless. While no final explanation for this phenomenon could be
provided, the assumption is, that the introduction of more short-term
dependencies enables the network to learn significant features. Hence, we apply this method to our proposed model. 

\subsection{Training Loss}\label{subsec:train_loss}
Autoencoders are usually trained with the mean squared error (MSE) as loss
function. The values of input and output can
therefore be compared in a value-by-value fashion. Nonetheless MSE is a general
metric that doesn't incorporate domain knowledge in its error computation. If the temporal dimension is only implicitly available through a fixed sampling rate, solely the reconstruction of the spatial information is important. In this case, the Euclidean distance between a point and its reconstruction suffices as loss metric.

Trajectories fed to and reconstructed by a neural network are however
normalized. If the Euclidean loss is computed on normalized trajectories, the actual distances are hidden. Hence, the training step can be further improved by rescaling the original and reconstruction
before loss calculation. 

When working with GPS data, trajectories are represented in geo-coordinates annotated with a timestamp. The temporal information is therefore explicitly available. Hence, an error calculation based on the Euclidean distance would be inaccurate. In this case, the haversine distance would be more general, more
accurate and more expressive. However, the computation is way more expensive than the Euclidean distance, while the equirectangular approximation offers a trade-off between both. Therefore, we consider this
approximation to calculate the distance between expected and predicted points. To incorporate the time in the loss function, the squared error of this dimension is added to the loss.

\vspace{-1.5em}
\section{Evaluation}\label{sec:eval}

\subsection{Datasets and Preprocessing}
For evaluation, we consider two datasets consisting of multiple recorded trajectories but varying in structure and origin as follows: 

\textbf{Quake III Dataset} consists of trajectories recorded in Quake III
Arena, a popular first-person shooter video game \cite{Breining2011}.
The player is able to move inside a fixed three dimensional area producing a series of three-dimensional positions. This location data is recorded at each position update which takes place every 33 ms - based on an frame rate of 30 fps. The
sampling rate here is therefore far higher than in real life scenarios. Furthermore, and in contrast to real positioning systems, there is no positioning error in the recorded location data. Temporal information is represented only implicitly through
the fixed sampling rate and accessible relatively to the start of the trajectory. The data was recorded in different scenarios and settings. Each run represents a trajectory from spawn to death. Since this dataset is recorded directly from the
game, the level of accuracy is high and there is no need for further preprocessing steps. 

\textbf{T-Drive} The original T-Drive dataset \cite{Yuan2010} contains recorded trajectories of about 33,000 taxis in Beijing over the course of
three months. We use the public part containing more than 10,000 taxi trajectories over the course of one week resulting in 15 million sample points. Each point contains a taxi ID, a timestamp and the position of the car in latitude/longitude representation. The duration between two samples lasts from few seconds to multiple hours. Due to inaccurate or faulty GPS-samples, there are inconsistencies in this dataset which should be removed in a preprocessing step. Hence, we only consider trajectories which are located in the metropolitan area of Beijing and show continuously increasing timestamps. Idle times should also be removed from the data since they add no informational value. Based on speed information, we also remove outliers that are probably introduced by technical errors. Hence, obvious inconsistencies are removed by these preprocessing steps, while the level of inaccuracy due to technical limitations stays realistic. 

\subsection{Training and Normalization}
Both datasets are split into two independent parts: 90\% of the recorded points are used for training and the remaining 10\% are used to test and validate the trained model to avoid overfitting effects. Our trajectories consist of varying numbers of time steps. Hence, we consider different sequence lengths of trajectories, such as $|T| = 20$ and $|T| = 40$.

In order to increase the amount of training data, a sliding window approach is
applied. If a trajectory consists of more time steps than
the required sequence length $|T|$, the first $|T|$ points $(P_{1}, ...,
P_{|T|})$ are added to the training data as input sequence. Afterwards, the
sliding window is shifted by one so that the steps $(P_{2}, ..., P_{|T|+1})$
build the next input sequence. This procedure is subsequently repeated until
the end of the trajectory is reached. Through this approach, the number of
sequences to train the model can be enlarged. The trajectories used for testing are not further manipulated but also split into chunks of 20 and 40 steps.

Similarities between trajectories are often based on the form of
the route they describe, rather than their absolute position in space. Rescaling each trajectory to values between 0 and 1 takes complexity out of the network leading to better performance and shorter learning time. For each
dimension, the minimum value (offset) and the scaling factor of a trajectory
are saved. The compressed trajectory therefore consists of the output of the
encoder network and the information to rescale the trajectory to its original
domain. After decoding the latent representation, the trajectory can be put to
its absolute position with those rescaling and offset values. Thus, the autoencoder has to learn the relative distances between the points of the
trajectory, rather than absolute values.

\subsection{Implementation and Setup}
The encoder was implemented as a plain LSTM cell, taking as input a mini-batch with a fixed number of time steps.
This decoding LSTM cell has an output dimensionality
of 100 in order to increase the ability to learn hidden features. 
As optimizer, we use Adam \cite{Kingma2014} at default parameters. 

Since the sequence length of the input trajectories is fixed for
the autoencoder, it should be evaluated, how well the model performs with
different given lengths $|T|$. Therefore, each experiment is performed once
with trajectories containing 20 \& 40 data points. 
Another
important property that has to be tested is the performance given different
compression ratios (c.p.~Table~\ref{table:scenarios_40}).
For the autoencoder the number of compressed values is the sum of the dimensionality of the latent variable and the variables needed for offset and rescaling.
For Douglas-Peucker it is the product of the number of points of the compression and their dimensionality.


Comparable compression ratios are limited by two factors.
First, Douglas-Peucker selects a subset of the original trajectory.
Hence, the amount of values in the compressed trajectory has to be divisible by 3.
Second, the proposed autoencoder model transfers the rescale variables together with the latent variable. Hence, the number of values in the compressed trajectory has to be greater than 6.

\begin{table}[tbh]
\centering
\resizebox{\linewidth}{!}{
	\begin{tabular}{cccc}
		\toprule
		Comp. Ratio & Autoencoder & Douglas-Peucker & Comp.
		Values \\
		\midrule
		2 & 54 + 6 & 20 $\cdot$ 3 & 60\\
		4 & 24 + 6 & 10 $\cdot$ 3 & 30\\
		8 & 9 + 6 & 5 $\cdot$ 3 & 15\\
		10 & 6 + 6 & 3 $\cdot$ 3 & 12\\
		\bottomrule
	\end{tabular}}
	\caption{Scenarios evaluated for sequence length 40.}
	\label{table:scenarios_40}
\end{table}

The errors between the original
and the compressed/reconstructed trajectories are subsequently calculated for
the three considered metrics, i.e., mean Euclidean/Haversine distance, Fr\'echet distance, and DTW. In the following sections, we present the obtained results for both datasets, separately.

\subsection{Performance on Quake III Dataset}\label{sec:results_quake}
First, we take the Quake III dataset.\\
\textbf{Trajectories of Sequence Length 20}
The autoencoder reconstructs the intermediate compression back to the original sequence length of 20 points,
while Douglas-Peucker yields 10, 5 and 3 points, respectively. For the
autoencoder reconstruction the mean Euclidean distance between points can be calculated intuitively, while the trajectories that are compressed
with Douglas-Peucker have to be interpolated first in order to equalize the
sequence lengths. Therefore, removed points are approximated on the line segments of the compressed trajectory by linear
interpolation. 


We notice, that the autoencoder performs worse than Douglas-Peucker for low compression ratio, which is the opossite for higher compression ratios. 
We assume, this is due to the way Douglas-Peuker selects a subset of the original trajectory:
For a compression ratio of 2, half of the points in the compressed
trajectory stay exactly the same.
For higher compression ratios, less points are identical and the distance becomes higher.
The difference between Douglas-Peucker and the autoencoder is all-in-all small and the tendency over different compression ratios is similar.
It has to be noted however, that the approximated locations can't be interpolated in this way in a real-life scenario, since the number of deleted points between the compressed locations can only be calculated if the original trajectory is still available.
The autoencoder model on the other hand reconstructs each point of th
trajectory.
Therefore, the comparison method favors Douglas-Peucker, but the autoencoder model still performs better, at least for higher compression ratios within our settings.

When analyzing the mean discrete Fr\'echet distances, the autoencoder performs significantly better than Douglas-Peucker across all compression ratios. However, sincethis metric is an approximation of the actual Fr\'echet distance, it is sensitive
regarding differences in the number of vertices of the compared polylines. 
This can also be verified by interpolating the deleted points.


We did also compare the interpolated autoencoder reconstructions to trajecetories, compressed with Douglas-Peucker.
While the autoencoder performs slightly better in terms of Fr\'echet distance for a compression ratio of 2, it performs worse for the other two ratios.
However, the compressed trajectories yielded by Douglas-Peucker had to be
enriched with information from the original trajectory, whereas the
reconstructions from the autoencoder were not manipulated. 

The trajectories considered here consist of three-dimensional
spatial positions. Hence, the consider DTW$_D$, since the dimensions are dependent of each other. As explained previously, DTW has similarities to the Fr\'echet distance and therefore it is sensitive to different sequence lengths. 
We notice similar tendencies in comparison to the discrete Fr\'echet errors from above. 
The distances regarding the autoencoder reconstructions are smaller than in case of Douglas-Peucker compressions without interpolated points. 

\textbf{Trajectories of Sequence Length 40} We evaluate trajectories with a sequence length of 40 positions. The higher length also allows higher compression ratios, e.g., 8
and 10. The corresponding results are shown in Table \ref{table:quake_euc_40} considering the mean Euclidean distances as error. 

\begin{table}[t!]
\centering
	\begin{tabularx}{\linewidth}{ccc}
		\toprule
		Comp. Ratio & Douglas-Peucker & Autoencoder\\
		\midrule
		2 & 1.45 & 1.76\\
		4 & 3.82 & 3.75\\
		8 & 14.85 & 13.88\\
		10 & 24.87 & 24.39\\
		\bottomrule
	\end{tabularx}
	\caption{Mean Euclidean distance for sequence length 40 with Quake III
	dataset.}
	\label{table:quake_euc_40}
\end{table}
It can be seen, that our previous observations can be replicated for longer trajectories. The autoencoder model performs worse for a
compression to half of the original size, while higher compression ratios
produce similar results with a tendency in favor of the autoencoder. That
implies that the temporal dependencies between the points can be captured for
both sequence lengths equally well.

In case of Fr\'echet and DTW, the results are also comparable to the lower sequence lengths, as shown in Table \ref{table:quake_frechet_40}. 
\begin{table}[t!]
	\centering
    \resizebox{\linewidth}{!}{
	\begin{tabular}{ l l r r r }
		\toprule
		& & \multicolumn{2}{c}{Douglas-Peucker} \\ \cmidrule{3-4}
		CR & & Normal & Interpolated & Autoencoder \\ \midrule
		\multirow{2}{*}{2} & Fr\'echet & 30.23 & 7.89 & 7.29 \\
		& DTW & 278.40 & 46.33 & 70.09 \\ \midrule
		\multirow{2}{*}{4} & Fr\'echet & 52.45 & 14.92 & 16.86 \\
		& DTW & 616.59 & 117.36 & 144.79 \\ \midrule
		\multirow{2}{*}{8} & Fr\'echet & 93.36 & 41.03 & 64.98 \\
		& DTW & 1304.55 & 423.59 & 509.51 \\ \midrule
		\multirow{2}{*}{10} & Fr\'echet & 114.63 & 58.32 & 87.33 \\
		& DTW & 1701.29 & 701.62 & 882.43 \\
		\bottomrule
	\end{tabular}}
	\caption[Mean Fr\'echet and DTW distances for sequence length 40 with
	Quake III.]{Mean Fr\'echet and DTW distances for sequence length 40 with
	Quake III dataset.}
	\label{table:quake_frechet_40}
\end{table}
For a better comparison, the trajectories compressed by Douglas-Peucker were
again interpolated to the same number of points as the original trajectories. The
distances between the originals and the reconstructions from the autoencoder
are far smaller than the ones between original and Douglas-Peucker
compressions. However, the distances between original and interpolated
compressions are lower than the autoencoder errors. Hence, we state that the main advantage of the autoencoder is the
reconstruction of every point of the original trajectory in contrast to the
selection of a subset. 

\begin{figure}[!t]
	\centering
		\subfloat[Mean Fr\'echet distance.]{%
			\includegraphics[width=.25\textwidth]{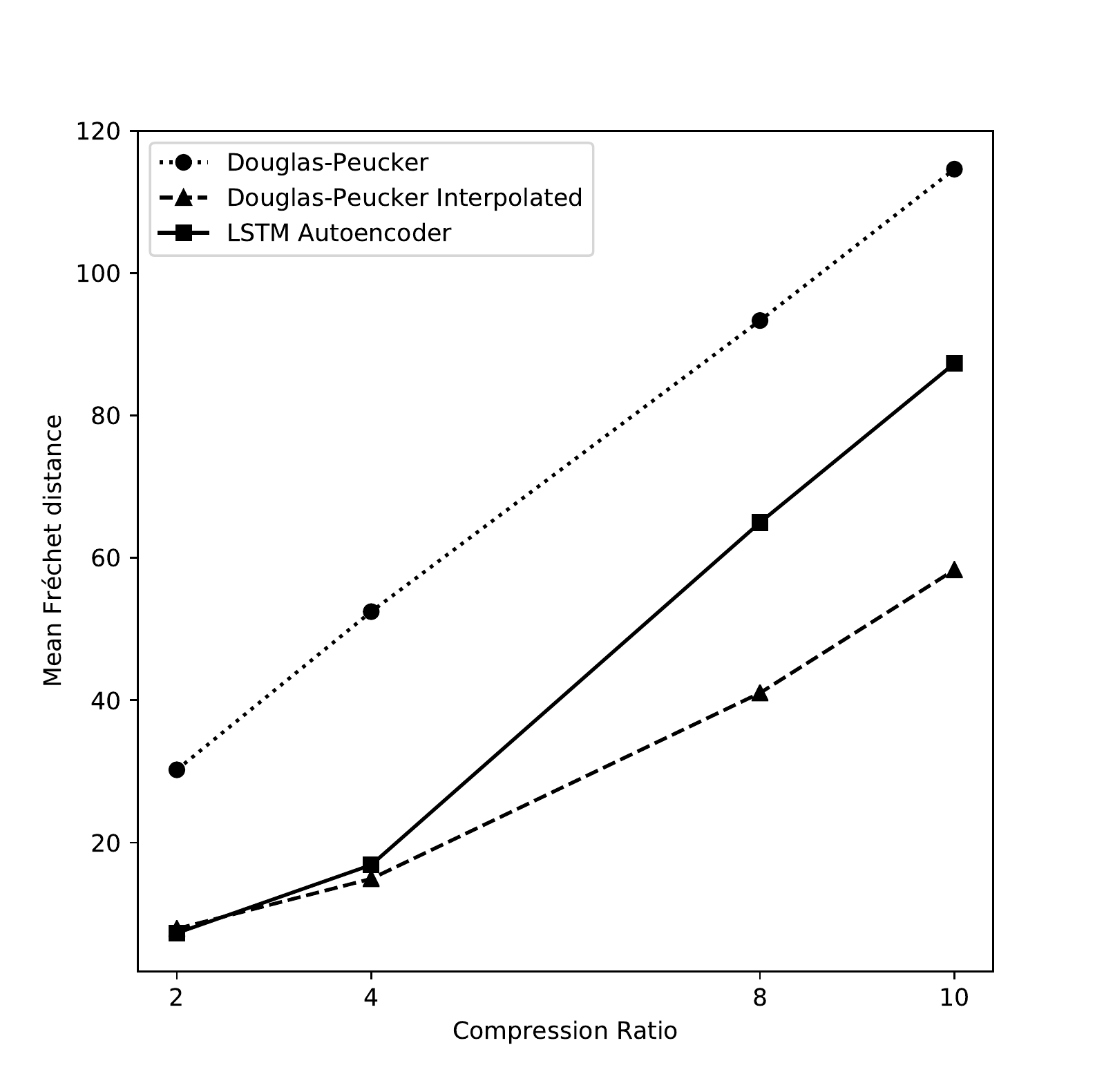}%
			\label{fig:frechet_40_plot}%
		}
		\subfloat[Mean DTW distance.]{%
			\includegraphics[width=.25\textwidth]{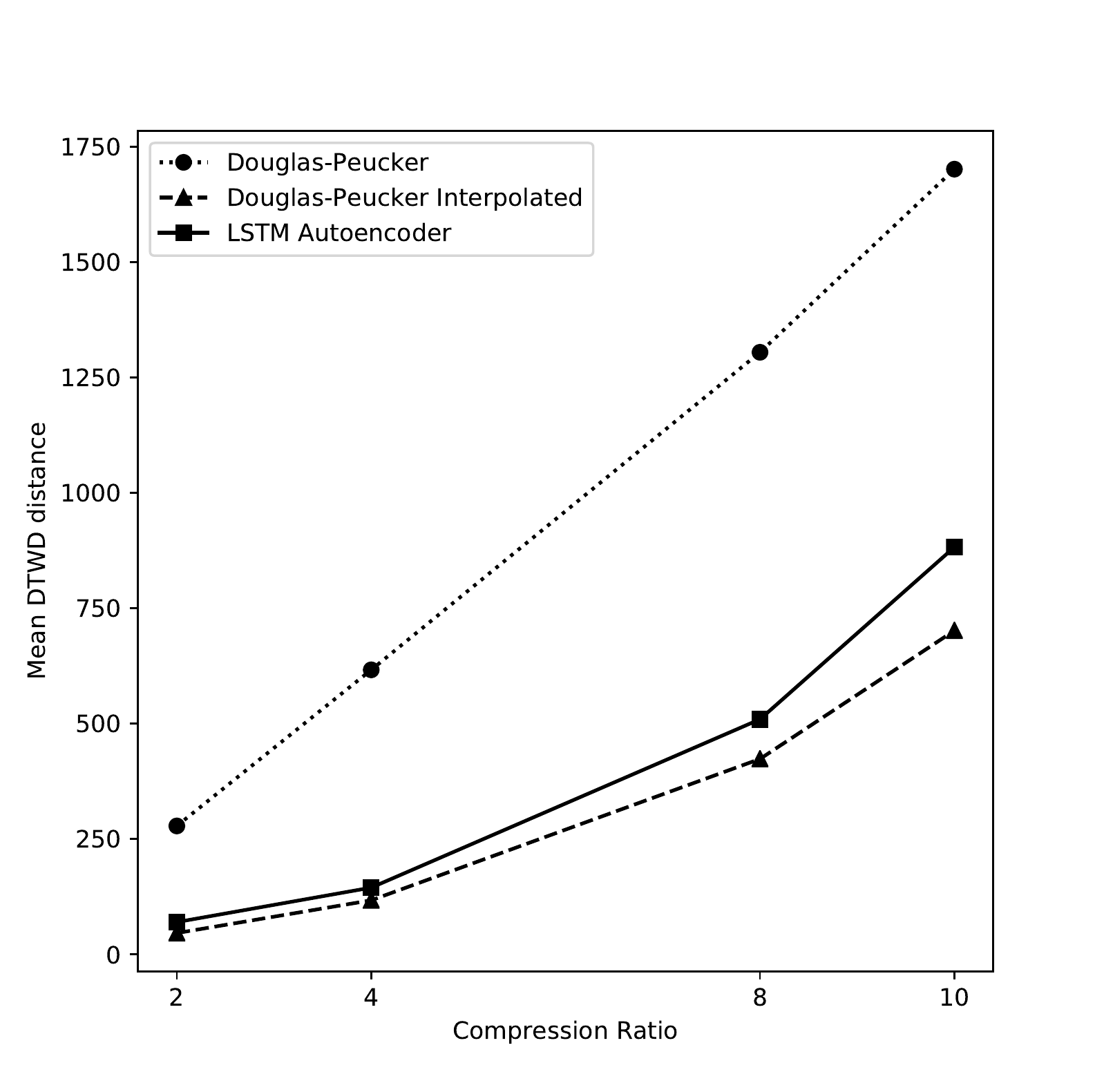}%
			\label{fig:dtwd_40_plot}%
		}
		\caption{Mean distances for sequence length 40 and compression
		ratios 2, 4, 8, 10.}
		\label{fig:distances_40_plots}
\end{figure}

Figure \ref{fig:distances_40_plots} shows the mean
Fr\'echet and DTW distances for a sequence length of 40 over the four tested
compression ratios.
The distances between original trajectories and their subsets produced by
Douglas-Peucker increase linearly over the compression ratio for both distance
metrics. The Fr\'echet distance to the autoencoder reconstruction is less than the distance between original and interpolated Douglas-Peucker compression for a ratio of 2, but for higher
ratios the autoencoder error grows faster. As explained before, the Fr\'echet distance is sensitive to outliers, since it
represents the maximum of the minimal distances between points. As the
distance grows faster for the autoencoder, more outliers are produced. This observation is supported by Figure \ref{fig:dtwd_40_plot} where the mean DTW distances are shown. The autoencoder distance here is always higher than the interpolated Douglas-Peucker distance, but the difference
between both methods grows slower for higher compression ratios compared to
the Fr\'echet distance in Figure \ref{fig:frechet_40_plot}. Since DTW measures the sum of minimal distances between points, this metric is not as sensitive to outliers as the
Fr\'echet distance. An explanation for the increased errors in case of higher compression ratios is the dimensionality of the latent space. With a trajectory of 40 three-dimensional
points and a compression ratio of 10, the compression contains only 12 values. Six values are already reserved for rescaling and offset, so that the
latent space has a low dimensionality of 6. Hence, it's hard to capture all temporal
dependencies.

\subsection{Performance on T-Drive Dataset}
In contrast to the Quake trajectories, the T-Drive data points consist of two-dimensional GPS locations and a temporal annotation. Since the spatial positions are now represented as
latitude/longitude pairs, the haversine distance is used instead of the
Euclidean distance and the error is expressed in meters rather than using abstract distances like before. In the following, we evaluate a spatial and a time-synchronized comparison separately. During the first, we omit temporal dimension, while for the second, we incorporate temporal information by synchronizing the spatial points in time before performing distance computations. 

\subsubsection{Spatial Comparison}
In order to see whether the results observed in the compression of the Quake
III dataset can be reproduced for real-life data, the temporal component of the
T-Drive dataset is not considered. However, the compression is done on the whole three-dimensional trajectories. As in the previous evaluation, the trajectories compressed with Douglas-Peucker are once
considered directly as a subset of the original data and once with the
interpolated approximations of the deleted points.


We notice, a slighty higher mean haversine error of the autoencoder than the mean distance to the interpolated compressions of Douglas-Peucker. This is contrary to the results obtained with 20-point trajectories from the Quake III dataset. However, the difference between both compression methods stays stable for all three
tested compression ratios. Like in the previous experiments, the discrete Fr\'echet
and DTW$_D$ error of the autoencoder is significantly lower than the error between
original trajectories and the subsets produced by TD-TR. The DTW$_D$ error of the
autoencoder is however higher than the error of interpolated TD-TR
compressions. The tendency that the autoencoder performs better than the
interpolated TD-TR compressions for lower compression ratios in terms of
Fr\'echet distance can be replicated for the T-Drive dataset. Overall, the results are similar and the performance of our model is not dependent on the compressed
datasets. 

Figure \ref{fig:tdrive_40_errors} shows the obtained errors for the
compression with sequence length 40. The results are similar to the
observations with 20 points per trajectory. This gives again reason for the
assertion that the model performs equally for both sequence lengths.
Furthermore, the results show that the difference between autoencoder and TD-TR
are smaller for higher compression ratios. This is especially visible for the
mean haversine distance, where the autoencoder performs better with compression
ratios of 8 and 10. 

\subsubsection{Time-Synchronized Comparison}
Since the trajectories compressed with TD-TR are a subset of the
original trajectories, the remaining points are already synchronized in time.
The interpolated compressions and the autoencoder reconstructions however
contain positions that are not synchronized and therefore do not consider the
temporal information. In order to include this additional dimension in the
evaluation, both interpolated TD-TR compressions and autoencoder
reconstructions are now synchronized before error calculation. 


We notice, that the performance of both compression methods differ to our previous observations. For TD-TR, the errors are lower if the
interpolated trajectory is synchronized in time. This is due to the usage of SED as metric to decide which points are
stored in the compressed version. Considering the autoencoder, the distances to time-synchronized reconstructions are higher, especially for low compression ratios. This indicates, that the accuracy of the reconstructed temporal dimension is low and hence, the positions of the
synchronized points are further away from the original positions. An adapted loss function may improve the accuracy, e.g., by additional weighting the reconstruction error of the timestamp, or calculating a time-synchronized distance.

\begin{figure}[hpbt]
	\centering
		\subfloat[Mean haversine distance.]{%
			\includegraphics[width=.49\textwidth]{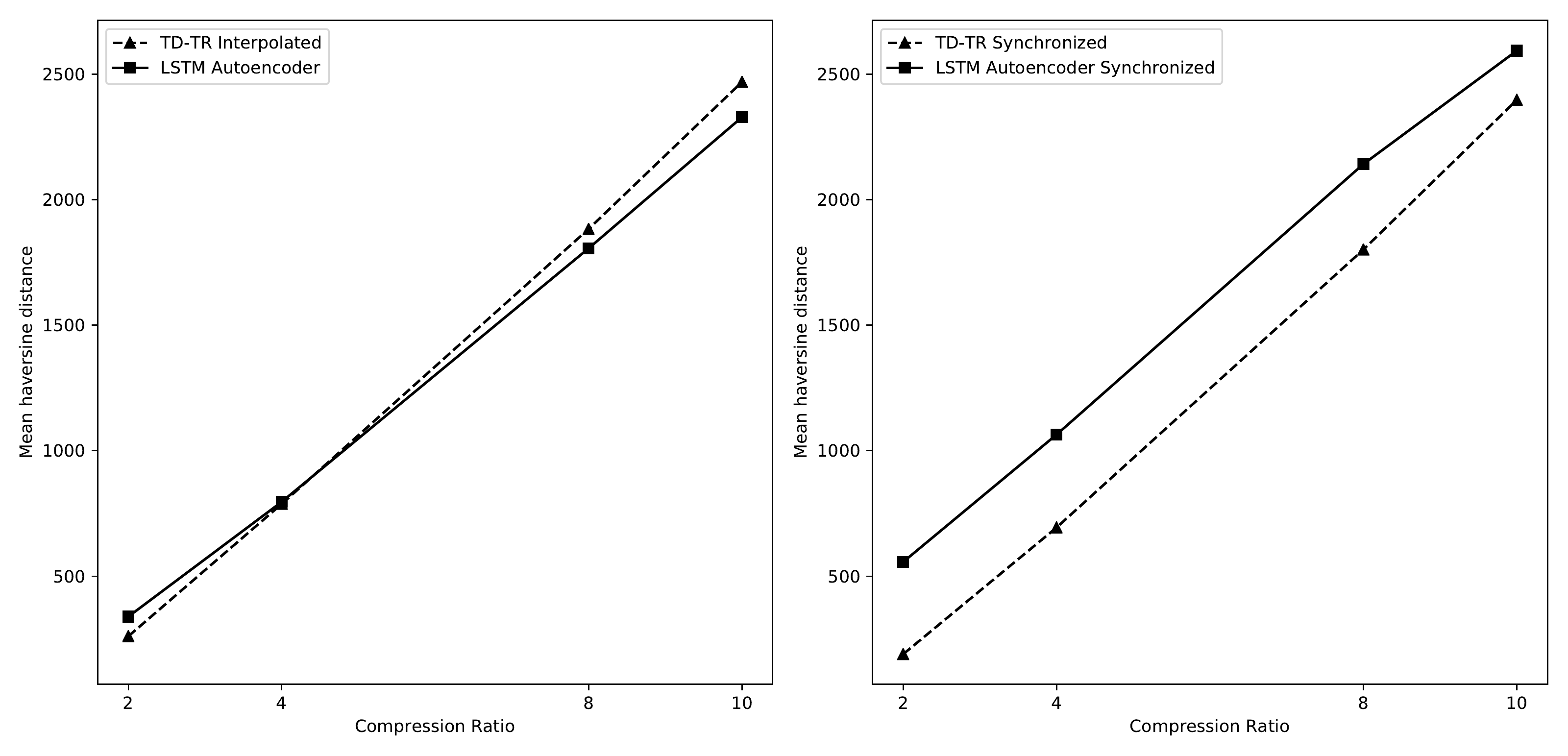}%
			\label{fig:haversine_40_plot_tdrive}%
		}\\
		\subfloat[Mean Fr\'echet distance.]{%
			\includegraphics[width=.49\textwidth]{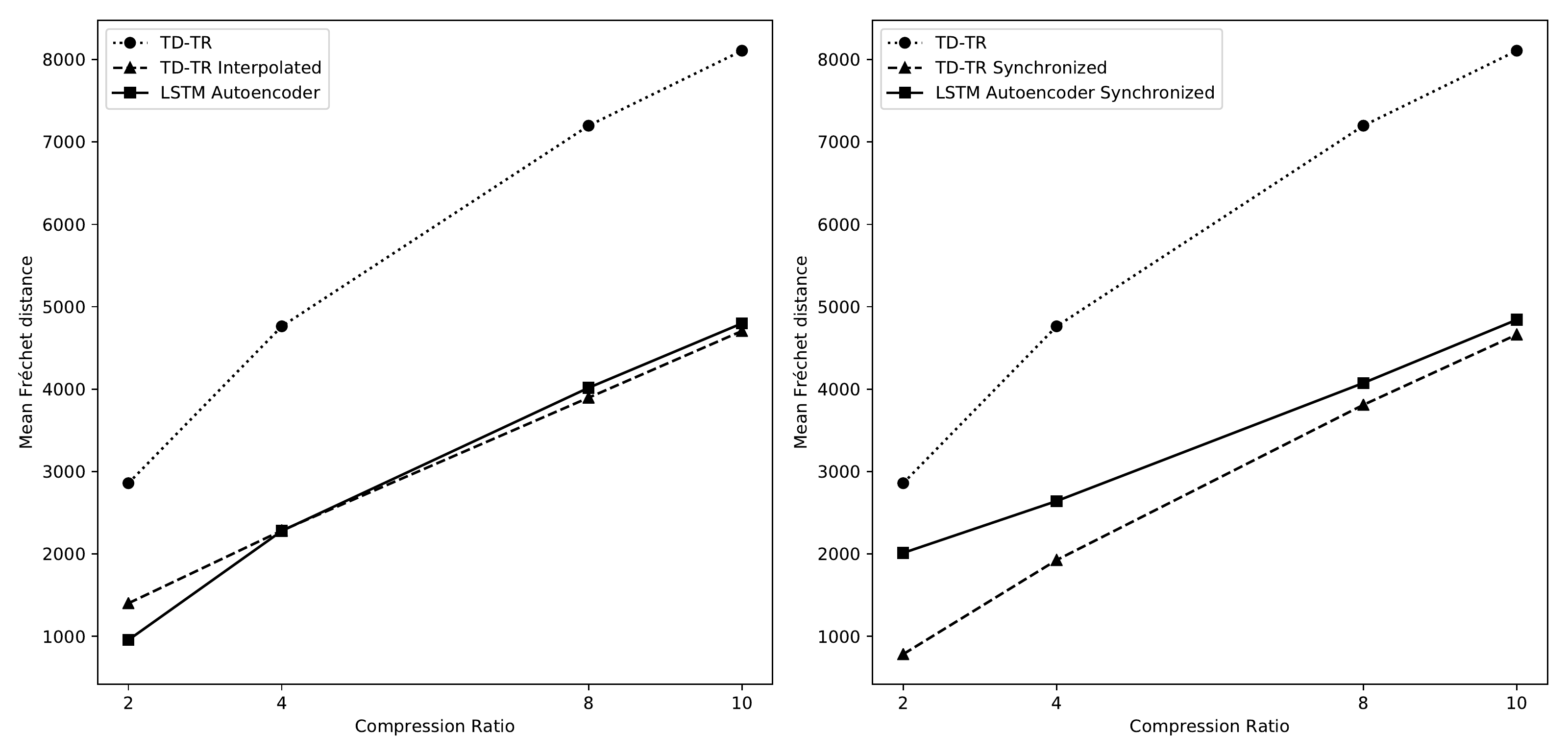}%
			\label{fig:frechet_40_plot_tdrive}%
		}\\
		\subfloat[Mean DTW distance.]{%
			\includegraphics[width=.49\textwidth]{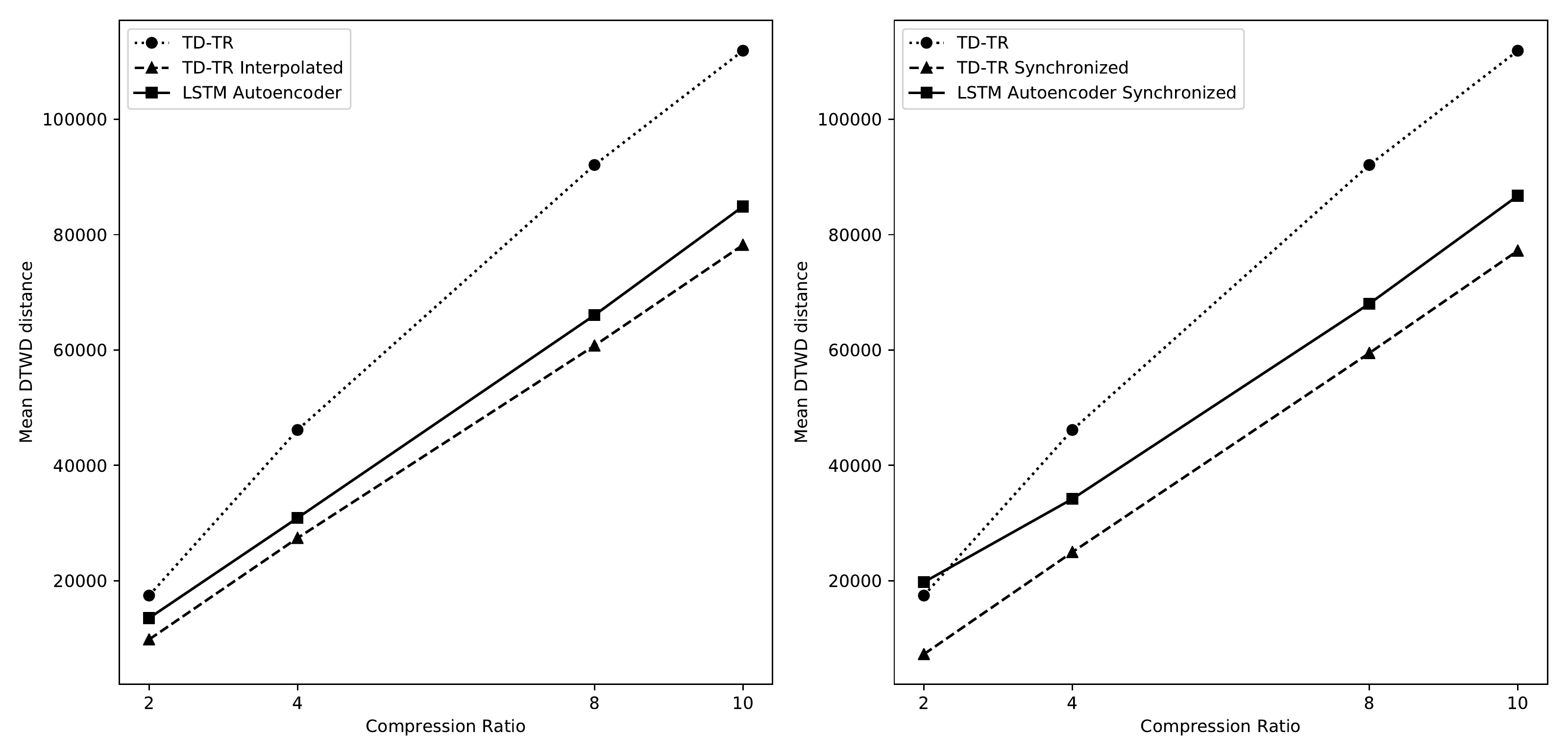}%
			\label{fig:dtwd_40_plot_tdrive}%
		}
		\caption[Errors of T-Drive dataset for sequence length 40 with
		time-synchronization.]{Mean distances for sequence length 40 and compression
		ratios 2, 4, 8 and 10 on the T-Drive dataset.
		Trajectories have been time synchronized on the right side.}
		\label{fig:tdrive_40_errors}
\end{figure}

The results for the time-synchronized error with a sequence length of 40 are
shown in the right column of Figure \ref{fig:tdrive_40_errors}. Obviously, the results from the 20-point trajectories and the spatial distances can be
reproduced. The autoencoder performs generally worse than the synchronized
TD-TR compression, but the difference is smaller for higher compression ratios.

\section{Conclusion and Future Work}\label{sec:FutureWork}
In this paper, we have investigated the feasibility of an autoencoder
approach for GPS trajectory compression. For this purpose, an LSTM network architecture was developed and evaluated using two diverse datasets. 

The obtained results lead to the conclusion that reconstructions of the proposed model outperform Douglas-Peucker compressions significantly in terms of discrete Fr\'echet distance and DTW
distance. With equalized sequence lengths through interpolation of the
Douglas-Peucker compressions, our autoencoder still produces a lower Fr\'echet
error for low compression ratios. For high compression ratios, the proposed
method achieves better results in terms of mean point-to-point distances depending on the representation space of the trajectories. While Douglas-Peucker selects a subset of the original
trajectory, the autoencoder approach reconstructs every point lossy. This
conceptual difference between the two methods is an advantage for the
autoencoder if temporal information of the trajectory is only available
implicitly through a fixed sampling rate. Furthermore, the reconstruction
approach allows a better retention of form and shape of the trajectory. The
performance of the autoencoder is reproducible for different trajectory lengths
as well as for different datasets. In contrast to most line simplification
algorithms, the user may fix the compression ratio before compression, which is advantageous for use cases where fixed storage or transmission sizes are expected. However, in terms of DTW distance, the autoencoder approach always
performed worse than Douglas-Peucker if equalized sequence lengths are
considered. Furthermore, the performance was worse for
time-synchronized trajectories. 

We summarize, that compressing GPS trajectories using an autoencoder model is feasible and promising. The performance of our proposed model could still be improved by various approaches. First, performing map-matching methods after decoding may decrease the reconstruction error, if an underlying road network is present on the decoder side. Second, the
usage of a stacked autoencoder which feeds the LSTM outputs at each time step to another LSTM layer, so that temporal and spatial dependencies are
captured more effectively. Generally, the performance can also be increased by using optimized hyperparameters and more efforts in training. Lately variational and adversarial autoencoder have proven to be successful advancements of normal autoencoders. Furthermore, the usage of a more efficient and differentiable loss function focusing on time series, such as Soft-DTW \cite{cuturi2017soft}, seems promising to improve our model significantly. 

In future work, we will enhance our investigations regarding these approaches. Overall, further research in this domain is highly promising and the incorporation of neural networks for trajectory compression could
create a new category next to line simplification and road network-based solutions.



\bibliographystyle{apalike}
{\small
\bibliography{main}}

\end{document}